# Cognitive Amplifier for Internet of Things

BING HUANG, University of Sydney, Australia

ATHMAN BOUGUETTAYA, University of Sydney

AZADEH GHARI NEIAT, University of Sydney

We present a *Cognitive Amplifier* framework to augment things part of an IoT, with cognitive capabilities for the purpose of improving life *convenience*. Specifically, the Cognitive Amplifier consists of knowledge discovery and prediction components. The knowledge discovery component focuses on finding natural activity patterns considering their *regularity*, *variations*, and *transitions* in real life setting. The prediction component takes the discovered knowledge as the base for inferring what, when, and where the next activity will happen. Experimental results on real-life data validate the feasibility and applicability of the proposed approach.

## 1 INTRODUCTION

With the emergence of IoT, there is a rising interest in applying Internet of Things (IoT) technology in the smart homes for making occupants' life more *convenient*. The *convenience* is underpinned by the principle of the *least effort*, i.e. the premise that humans would usually want to achieve goals with the least cognitive and physical efforts [2]. IoT refers to the networked interconnection of everyday things, which are augmented with capabilities such as sensing, actuating, and communication [21]. The availability of IoT devices including switch sensors, infrared motion sensors, pressure sensor, wearable sensors, accelerators, temperature, humidity, and light sensors have the potential to realize the convenience. It is a challenge that IoT devices are highly diverse in supporting infrastructure such as different programming language and communication protocols [5]. Service-oriented computing is a promising paradigm for abstracting connected things as services through hiding the complicated infrastructure [6]. The service paradigm provides a high level of abstraction, shifting our focus from infrastructure to how services are to be used. Connected things are referred as *IoT services*. Thus, common things such as a connected air-conditioner can be represented as an air-conditioner service.

There has been a large body of research conducted in real-life testbeds such as CASAS project, MavHome project, Gator Tech Smart House, iDorm, and Aware Home [1]. Current research on IoT mainly focuses on augmenting IoT services with basic capabilities such as sensing, actuating, and communication. The IoT services can provide limited convenience for the occupants. For example, when an occupant appears in the bedroom, the light is turned on automatically. The light is turned off after a period whether is no motion detected. The aforementioned research enables IoT services to see, hear, and smell the physical world [7], which is indeed a

---

Authors' addresses: Bing Huang, University of Sydney , School of Computer Science, Sydney, NSW, Australia, bhua6675@uni.sydney.edu.au; Athman Bouguettaya, University of Sydney , School of Computer Science, Sydney, NSW; Azadeh Ghari Neiat, University of Sydney , School of Computer Science, Sydney, NSW.

---

| IoT service event sequence | | | Periodic probabilistic composition patterns |
|---|---|---|---|
| Microwave | 9:59:56 | 10:00:03 | Preparing breakfast |
| Stereo | 10:00:49 | 10:14:10 | (6:00-7:00, in the kitchen) |
| Freezer | 10:03:13 | 11:36:16 | |
| Sink faucet | 11:30:09 | | Taking a shower |
| | | 11:30:21 | |
| | ……… | | (22:00-23:00, in the bathroom) |
| Refrigerator | 18:09:34 | 18:09:41 | ……….. |
| Cabinet | 18:11:19 | 18:11:32 | Going to bed |
| Toaster | 18:11:54 | 18:11:56 | (23:00-0:00, in the bedroom) |
| Door | 18:12:21 | 18:12:26 | |
| Composition instances | | | Temporal relationships among composition patterns |
| Preparing lunch | 12:30-12:45 | | |
| Washing dishes | | | (Preparing lunch) *before* (Washing dishes) |
| | 13:20-13:40 | | |
| Listening to music | 13:10-14:00 | | (Washing dishes) *during* (Listening to music) |

Fig. 1. Examples of the discovered knowledge

fundamental research topic. We argue that these basic capabilities are not sufficient to fully realize IoT services potentials. IoT services should be augmented with more sophisticated capabilities, i.e., *cognitive* capabilities. Similar to humans, the IoT devices can not only see, hear, and smell the physical world, but can also learn and understand the physical world. We consider the cognitive capability and empower IoT services with such high-level intelligence to fully fulfill IoT services' potential. Specifically, we develop a *Cognitive Amplifier* based on IoT paradigm. *The Cognitive Amplifier is an intelligent system aiming at providing convenience through anticipating what people would like to do and helping people to do it with less effort and time.* The key tasks of fulfilling the cognition for the Cognitive Amplifier are to enable IoT services to understand the ongoing activities that the occupant is engaged in and predict what, when, and where the occupant will do next.

Many supervised activity prediction methods have been proposed to predict what an occupant will do next without considering when and where the activity will take place. Moreover, supervised methods are based on labeled data. Labeling human activity data is time-consuming and laborious. It is also limited in the system scalability. In this paper, we first discover *knowledge* regarding daily activity patterns from unlabeled data and then use this knowledge to build the *prediction model*. We also refer the activity patterns as *composition patterns*. Specifically, we use the Cognitive Amplifier as a framework to provide *quantifiable convenience* for occupants through the development of the prediction model based on the knowledge including *periodic probabilistic composition patterns* and *temporal relationships among composition patterns*. The Cognitive Amplifier is potential to liberate the occupants from the repetitive and cumbersome interactions with IoT services. To the best of our knowledge, little attention has been paid to the issue of developing cognitive capabilities for IoT services. The Cognitive Amplifier consists of two components, namely knowledge discovery and prediction components. For the knowledge discovery component, we focus on discovering two types of knowledge including *periodic probabilistic composition patterns* and *temporal relationships among composition patterns* as shown in Fig.1. The periodic probabilistic composition patterns focus on capturing the *variation* and *periodicity* features of composition patterns. For example, the occupant has the habit of preparing breakfast around 6am to 7am in the kitchen. The temporal relationships among composition patterns aim to model the *transition* feature among composition patterns. For instance, the occupant may have the habit of listening to music while washing dishes (i.e., (Washing dishes) *during* (listening to music) as shown in Fig.1). For the prediction component, a prediction model is developed to forecast what, when, and where the occupant will do next. A key novel ingredient of our approach is also the transformation of *qualitative convenience* to *quantitative convenience* as a key mechanism for assessing the optimality of cognitive amplifier. The key contributions are as follows:

- A new composition pattern is modelled to represent an activity. The quality of composition pattern is a key issue. We may discover numerous composition patterns. Many of them may be of low quality. We devise a quality model in terms of *significance* and *proximity* to discard low-quality composition patterns.
- A periodic probabilistic composition pattern model is proposed to describe the regularity and variation of the composition patterns. The periodic composition patterns are modelled by the spatio-temporal relationships among composition patterns by introducing an *involvement probabilistic concept* and *periodicity (i.e., a representative time interval and a location)*. The involvement probabilistic concept is used to capture the variations of the activities. The periodicity is also introduced to describe the regularity of the activities.
- A temporal relationship model is proposed to describe the transitions and relationships among composition patterns. The temporal relationships among composition patterns are captured in a matrix based on temporal logic.
- The discovered periodic probabilistic composition patterns and temporal relationships among composition patterns serve the knowledge base for constructing the prediction model of the Cognitive Amplifier. We propose the prediction model based on the integration of the two types of knowledge.
- A set of new algorithms is devised for the Cognitive Amplifier. We adapt the Prefixspan algorithm and K-means algorithm to efficiently discover periodic probabilistic composition patterns. A new algorithm TPMiner is also proposed to extract the temporal relationships among composition patterns. Finally, we propose a new algorithm to predict what, when, and where the occupant will do next.
- A new convenience model is proposed to quantify convenience in terms of *saving efforts* and *saving time*.

The rest of the paper is organized as follows. Section 2 presents the system model for the Cognitive Amplifier. Section 3 describes the periodic probabilistic composition patterns and the temporal relationships among composition patterns. Section 4 illustrates the prediction model based on the discovered knowledge. Section 5 elaborates the details of the knowledge discovery approach. Section 6 evaluates the approach and shows the experiment results. Section 7 presents the related work. Section 8 concludes the paper and highlights some future work. Motivating Scenario

The purpose of a smart home is to make occupants' daily life more convenient. Life *Convenience* can be seen as the benefits in terms of saving efforts and time, which is brought by an intelligent system (i.e., Cognitive Amplifier). Such a system is able to understand occupants' ongoing activities and react accordingly. This intelligent system has the potential to reduce occupants' interactions with IoT services. We consider an interesting scenario showing how IoT services bring about convenience for Sarah. On a Tuesday morning, the clock starts to ring at 8am. Meanwhile, the light in the bedroom is turned on automatically. Then, the Spotify app is playing popular songs to fresh her. At the same time, the heater is heating the bathroom. Sarah quickly gets up and goes to the bathroom to take a shower. When she leaves the bedroom, the light is turned off and the music is stopped automatically. The bathroom is already warm and she does not need to wait. When she is taking a shower, the music player in the bathroom is playing her favorite music. Meanwhile, the coffee maker is making coffee in the kitchen. After finishing showering, she goes to the kitchen. The coffee is ready. The TV in the living room is turned on automatically when Sarah has breakfast.

A key task to provide convenience is to enable the Cognitive Amplifier to understand IoT services usage patterns. The usage history of IoT services is recorded as IoT service event sequences. Fig.2. shows a three days IoT service event sequences. For example, $< B^+B^-,(5, 23)>$ describes the lamp is turned on at time 5 and turned off at time 23. $B^+$(resp.$B^-$) represents a turn on (resp. turn off) the lamp event. The set of spatio-temporal correlated IoT services illustrate an activity. As shown in Day 1, the relations between the shower (i.e., $F$) and the music player (i.e., $E$) shows that Sarah is taking a shower while the music player is playing music in the living room. As a result, a set of IoT services that are frequently used together in particular spatio-temporal relations may denote a daily activity. Such set of IoT services are termed as a *composition pattern*.

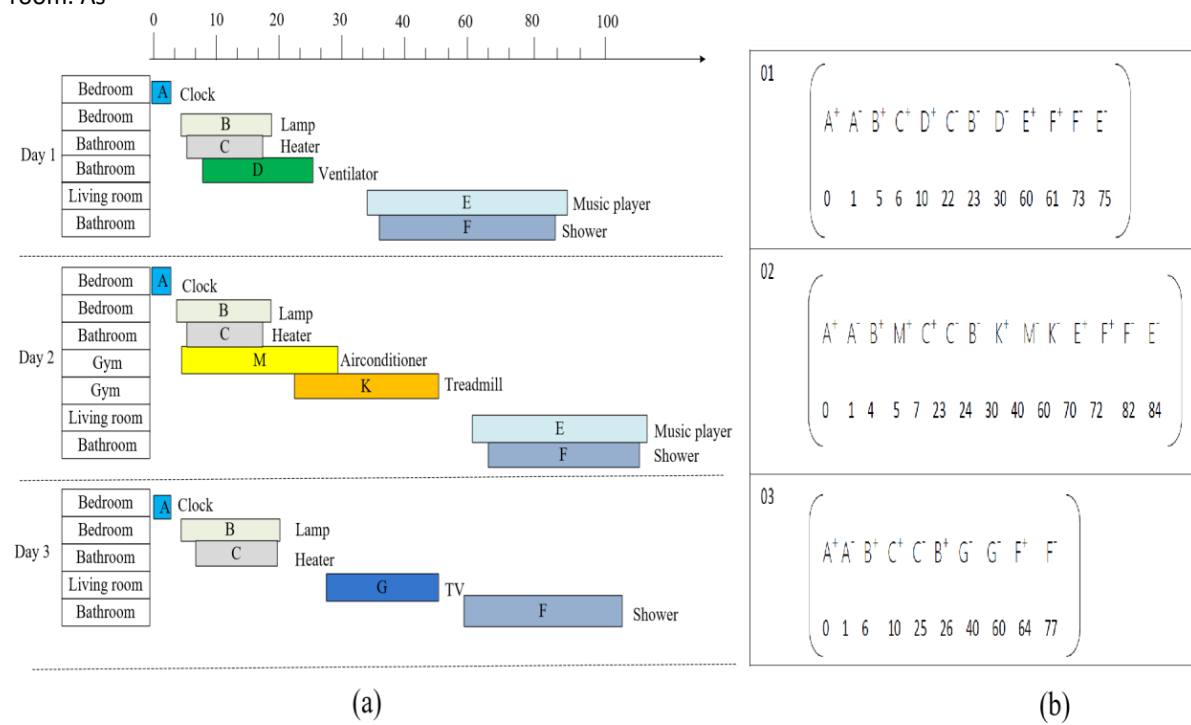

Fig. 2. An example of the IoT service event sequences

It is possible to establish numerous spatio-temporal correlations among IoT services, resulting in many composition patterns. Many of the composition patterns are of low quality because these correlations may be loose and insignificant. For example, in Day 1 as shown in Fig. 2(b), it is possible to generate many composition patterns through a brute force method. Thus, we propose the quality measure in terms of *significance* and *proximity* to discard these low-quality composition patterns.

We identify three key features for composition patterns. First, the composition pattern has a *variation* feature. Variation feature shows that the occupant sometimes may perform the activity differently. Some of the steps may be changed or varied in order and some IoT services may not be used in the same activity. Second, the composition pattern has a *periodic* feature. *A periodic composition pattern can be loosely defined as repeating composition patterns at certain locations with regular time intervals*. For instance, an occupant may have the habit of going to bed around 11 pm to 12 pm in the living room. We define the notion of *periodic probabilistic composition patterns* to capture both the periodicity and variation features of the composition patterns as shown in Fig.1. An example of this pattern is preparing breakfast at around 6:00-7:00 in the kitchen. Third, the composition pattern *transits* from one to another in a particular temporal relationship. The occupant may perform multiple activities in a sequentially or parallel manner. Whenever there are sequential and parallel activities, it may imply that there may be useful *temporal relationships* hidden among activities. For example, an occupant usually prepares lunch before wash dishes or they have a habit of washing dishes while listening to music as shown in Fig.1. In this regard, we propose the *temporal relationship model among composition patterns* to capture the *transition* feature of the composition patterns.

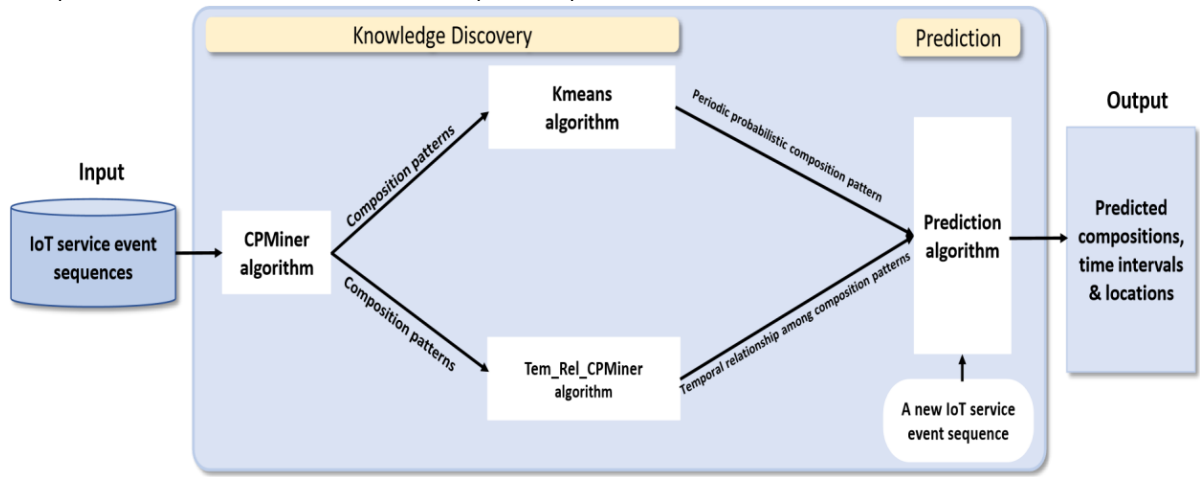

Fig. 3. The framework of the Cognitive Amplifier

The combination of these two types of knowledge including the periodic probabilistic composition patterns and temporal relationships among composition patterns provide the knowledge base for predicting future activities. Given a new IoT service event sequence, the prediction model can recognize the ongoing activities and predict what, when, and where the next activity will be taken place.

## 2 SYSTEM MODEL

A key goal of smart homes is to provide convenience for the occupants. In this section, we first propose an IoT service model and a new convenience model in terms of saving efforts and time. We then propose Cognitive Amplifier framework to achieve convenience.

Definition 1: IoT service . An IoT service $s_i$ is a tuple $s_i =< ID, F_i, IS, FS, L_i >$, where:

- $ID$ is a unique service identifier.
- $F_i$ is a set of functionalities offered by $s_i$ • $IS$ (Initial State) is a tuple $< s_i^+, st_i, sl_i >$, where – $s_i^+$ is a symbol of $IS$.
  - $st_i$ is a start-time of $S_i$.
  - $sl_i =< x_s, y_s >$ is a start location of $s_i$, where $< x_s, y_s >$ is a coordinate point.
- $FS$ (Final State) is a tuple $< s_i^-, et_i, el_i >$, where
  - $s_i^-$ is a symbol of $FS$
  - $et_i$ is an end-time of $s_i$.
  - $el_i =< x_e, y_e >$ is an end location of $s_i$, where $< x_e, y_e >$ is a coordinate point.
- $L_i$ (Location) is the semantic location of $s_i$.

In the rest of the paper, we focus on spatio-temporal aspects of IoT services. Therefore, the IoT service model can be simplified as $< (^{S_i^+}, st_i, sl_i), (^{S_i^-}, et_i, el_i), L_i >$. An example of a light service can be denoted as $< (L100^+, 8pm, (2, 3)), (L100^-, 10pm, (2, 3)), kitchen >$. It represents the light is using from 8pm to 10pm in the kitchen. The point $(2, 3)$ is the coordinate in the kitchen and 8pm (resp. 10pm) is the start time (resp. end time).

The execution records of IoT services are referred as an *IoT service event sequence*. For example, Fig. 2(b) shows an example of three IoT service event sequences.

After using Cognitive Amplifier for a period of time, we can quantify how much convenience is obtained. The convenience is defined as follows.

Definition 2: Convenience. Given a period of time $T$ and the IoT service event sequence during $T$ { $<(x_1^+, st_1, sl_1), (x_1^-, et_1, el_1), l_1 >...< (x_n^+, st_n, sl_n), (x_n^-, st_n, sl_n), l_n >$}, the Cognitive Amplifier infers the set of IoT service events { $< y_1, y_2...y_m >, t_y, l_y$ } that will occur in the next time segment $t_y$ and at the location $l_y$. Suppose an actual event set occurs next is { $c_1, c_2...c_k$ }. Therefore, the convenience is defined as a tuple *Convenience* $=< saving\_efforts, saving\_time >$. The $saving\_efforts$ reflects the amount of reduced efforts in terms of reducing interactions with IoT services, which is formalized in Equation (1).

$$saving\_efforts = \frac{|\{y_1, y_2...y_m\} \cap \{c_1, c_2...c_k\}|}{|\{c_1, c_2...c_k\}|} \quad (1)$$

where $|\{c_1, c_2...c_k\}|$ is the number of actual future events, $|\{y_1, y_2...y_m\}|$ is the set of predicted events, and $|\{y_1, y_2...y_m\} \cap \{c_1, c_2...c_k\}|$ is the number of correctly predicted events.

The $saving\_time$ measures the amount of saved time. In real life scenarios, some IoT services have waiting time, which means residents need to wait for a period of time until consuming the IoT services' functionality. For example, the resident usually needs to wait for some time (e.g., 5 minutes) to preheat the bathroom before taking a shower. The Cognitive Amplifier can help a resident to save time by reducing the waiting time. For example, it can help the occupant by turning on the heater ahead of 5 minutes to heat the bathroom. Thus the occupants do not need to wait for 5 minutes to take a shower. Suppose $\{z_1, z_2...z_h\} = \{y_1, y_2...y_m\} \cap \{c_1, c_2...c_k\}$ is the set of correctly predicted events. For each $z_i$, the occupant needs to wait for $wait_i$ ( $wait_i \geq 0$ ) amount of time to get the delivered IoT service functionality. The total amount of saved time can be computed by

$$saving\_time = \sum_{i=1}^{h} wait_i. \qquad (2)$$

The key task of fulfilling convenience is to predict what, when, and where the occupant will do next based on the ongoing activities. Specifically, the key task is to predict what, when, and where the event sequence will occur, denoted as $\{< y_1, y_2...y_m >, t_y, l_y\}$, given the ongoing event sequence $\{ <(x_1^+, st_1, sl_1), (x_1^-, et_1, el_1), l_1 >...< (x_n^+, st_n, sl_n), (x_n^-, st_n, sl_n), l_n >\}$. The Cognitive Amplifier is proposed as a framework to provide quantifiable convenience for occupants of smart homes. It consists of the knowledge discovery and prediction components. The knowledge discovery component focuses on extracting periodic probabilistic composition patterns and temporal relationships among the composition patterns from unlabeled event sequences. The prediction model focuses on forecasting what, when, and where the next set of service events will occur based on the knowledge discovery. The two components are detailed in Section 3 and 4.

## 3  KNOWLEDGE DISCOVERY

In this section, we describe periodic probabilistic composition patterns and temporal relationships among composition patterns.

### 3.1  Periodic Probabilistic Composition Patterns

We present a three-level periodic probabilistic composition pattern model to capture the occupants' daily activity patterns. We first introduce the notion *composition patterns* based on spatio-temporal features [8]. The composition pattern actually represents an activity. A *quality model* in terms of statistic *significance* and *proximity* are proposed to prune low-quality composition patterns. Then we introduce the new model *probabilistic composition pattern* to capture the variation of the composition patterns by introducing involvement probability concept. Lastly, we propose the novel notion *periodic probabilistic composition pattern* to capture the regularity of the composition patterns by introducing representative time intervals and locations. *3.1.1 Composition Pattern.* Usually, occupants collectively use multiple IoT services based on time and location correlations to complete a specific daily activity. We call such spatio-temporal correlated IoT services as a composition instance. Note that the composition instance is an initialization of the composition pattern. Definition 3: Composition pattern. A composition pattern *CP* is a set of IoT services that frequently occur together in a particular spatio-temporal relationship. A composition pattern is represented by a tuple *CP* =< *S,Seq,sup(Seq)* > where

- $S = \{S_1, S_2...S_{sup}\}$ is the set of composition instances where $S_k = \{< (s_1^+, st_1, sl_1), (s_1^-, et_1, el_1) >, ..., < (s_n^+, st_n, sl_n), (s_n^-, et_n, el_n) >\}$ is a composition instance and there are *sup* numbers of composition instances. Note that $< (s_i^+, st_i, sl_i), (s_i^-, et_i, el_i) >$ is a component IoT service of $S_k$ defined in Definition 1 and $st_i \leq st_{i+1}$ and $st_i \leq et_i$. We sort all elements $s_i^*$ (* can be + or -) in $S_k$ in an increasing order based on its associated time information $st_i$(or $et_i$). Then, $S_k$ is transformed into the following representation:

$$S_k = <Seq, T, L> = \begin{Bmatrix} \alpha_1 & ... & \alpha_i & ... & \overline{\alpha_{2n}} \\ t_1 & ... & t_i & ... & \overline{t_{2n}} \\ l_1 & ... & l_i & ... & l_{2n} \end{Bmatrix},$$

where $Seq = \{\alpha_1...\alpha_i...\alpha_{2n}\}$ is a symbol sequence and $\alpha_i = s_j^*$ ($*$ can be + or -), $T = \{t_1...t_i...t_{2n}\}$ is the time information and $t_i \leq t_{i+1}$, and $L = \{l_1...l_i...l_{2n}\}$ is the location information. For example, one composition instance (music player and shower in Day1) in Fig. 2 can be represented as follows.

$$\left\{\begin{matrix} E^+ & F^+ & F^- & E^- \\ 60 & 61 & 73 & 75 \\ l_1 & l_2 & l_2 & l_1 \end{matrix}\right\}$$

$l_1 = (1, 2)$ and $l_2 = (2, 4)$ are coordinates.

- $Seq = \{\alpha_1...\alpha_i...\alpha_{2n}\}$ is the frequent sequence shared by all composition instances in $S$. An example of frequent sequence $Seq$ in Fig. 2(b) is $< E^+F^+F^-E^- >$.
- $sup(Seq)$ is the support for $Seq$. The $sup(Seq)$ is the number of composition instance occurrence in $S$ where each composition instance contains $Seq$. Suppose $S_k = < Seq, T, L >$ and $S' = < Seq', T', L' >$ are two composition instances. $S_k$ is said to be contained by $S'$, denoted as $S_k \sqsubseteq S'$, if $Seq = \{\alpha_1...\alpha_i...\alpha_{2n}\}$ is a subset of $Seq' = \{\alpha'_1...\alpha'_i...\alpha'_{2m}\}$, represented as $Seq \sqsubseteq Seq'$ and $n \leq m$. $Seq \sqsubseteq Seq'$ is satisfied if there exist integers $1 \leq k_1 \leq k_2...k_n \leq k_{2m}$ such that $\alpha_1 \subseteq \alpha'_{k_1}$, $\alpha_2 \subseteq \alpha'_{k_2}$,..., $\alpha_n \subseteq \alpha'_{k_n}$.

An IoT service event sequence database DB is a set of tuples $(sid, S')$ (i.e., $sid$ is a sequence ID and $S'$ is the composition instance). The tuple $(sid, S')$ contains $S_k$ if $S_k \sqsubseteq S'$. The support $sup(Seq)$ of $Seq$ in DB, denoted as $sup(Seq)$ is the number of tuples containing $Seq$. The $sup(Seq)$ can be formalized as follows.

$$sup(Seq) = |\{(sid, S_k) \in DB | S_k \sqsubseteq S'\}| \qquad (3)$$

For instance, the support for $< E^+F^+F^-E^- >$ is 2. For the sake of clarity, we refer the frequent sequence $Seq$ as the composition pattern and the set $S$ as the supporting composition instances for $Seq$.

*3.1.2 Quality Model for Composition Patterns.* We may discover a large number of composition patterns by only considering their occurrence frequency and many of them may be of low quality. Thus, we design a quality model to discard these low-quality composition patterns based on statistical *significance* and *proximity* perspectives. On the one hand, it is possible that composition patterns with high occurrence frequency are more significant than those with low occurrence frequency. Therefore, we devise a *si∂nif icance* technique to compute how much significance is for composition patterns from a statistical view. On the other hand, it is obvious that IoT services occurring spatio-temporal closely are more probable to be related. For instance, from the spatial aspect, the TV and the light in the same dining room may be highly correlated. However, the TV in the living room and the light in the dinning room may not be correlated even they are used together. From the temporal aspect, the TV and the light may be correlated when they are used together during the night. However, the TV and the light may not have any relations if the TV is used during the night and the light is used in the morning. Hence, we propose a *proximity* model to characterize the average strength of correlation for composition patterns. The proximity model is designed from *spatial-proximity* and *temporal-proximity* aspects. The *spatial-proximity* models the average strength of spatial correlations among composition instances. The *temporal-proximity* models the average strength of temporal correlations among composition instances. The *proximity* model is adapted from the method for measuring the distance among spatio-temporal interval data [22]. Given a composition pattern $CP =< S, Seq, sup(Seq) >$, its quality is defined as a tuple *Quality =< si∂nif icance, proximity >* where the two aspects are formalized as follows.

Definition 4: Significance. *Si∂nif icance* is used for evaluating the statistical importance of the composition pattern $CP$. It is formalized as:

$$significance(CP) = \frac{\sqrt{expect(Seq)}}{sup(Seq) - expect(Seq)} \qquad (4)$$

where *expect*(*Seq*) is the expected number of composition patterns occurrence in the *DB*. We adapt the model proposed in [23] to estimate *expect*(*Seq*) by considering different IoT service usage frequency in different regions in smart homes. In fact, it is obvious that IoT service usage frequency is different in different regions [4]. For instance, an occupant may spend most of his/her time in the bedroom during the daytime and only goes to the bathroom to take a shower. Thus, IoT services in the bedroom are likely to be used more frequently than those in the bathroom. As a result, the composition pattern for taking a shower may be ignored when only the support/frequency measure is employed for searching composition patterns.

For a composition pattern $CP = < S, Seq, sup(Seq) >$, $Seq = \{\alpha_1...\alpha_i...\alpha_{2n}\}$ is a possible outcome drawn from the symbol event set $E = \{e_1...e_i...e_n\}$ with probability of $P(e_i)$ following Bernoulli distribution such that $\sum_{i=1}^{n} P(e_i) = 1$.

Given a $DB$ and a region set $R = \{r_1, ..., r_k\}$, $DB_{r_i}$ contains IoT service event sequences that occur at region $r_i$. $Num(e_i)_{DB_{r_i}}$ denotes the occurrence frequency of the event $e_i$ in $DB_{r_i}$. $P(e_i)$ is estimated by $\frac{Num(e_i)_{DB_{r_i}}}{\sum_{e_j \in E} Num(e_j)_{DB_{r_i}}}$. Thus, the occurrence probability $P(Seq)$ and the expected number of occurrence $expect(Seq)$ are defined as follows.

$$P(Seq) = \prod_{\forall \alpha_i \in Seq} P(\alpha_i) \tag{5}$$

$$expect(Seq) = P(Seq) \cdot |DB_{r_i}| \tag{6}$$

where $|DB_{r_i}|$ is total number of IoT service events in region $r_i$. Note that we only consider $P(\alpha_i)$ when $\alpha_i = s_i^+$ and ignore $\alpha_j$ when $\alpha_j = s_i^-$ because $< s_i^+, s_i^- >$ are the event pair. For example, the significance of the sequence $< E^+ F^+ F^- E^- >$ is 0.35.

**Definition 5: Proximity.** Given a composition pattern $CP = < S, Seq, sup(Seq) >$ where a component IoT service $S_k$ in $S$ is represented as:

$$S_k = < Seq, T, L > = \begin{Bmatrix} \alpha_1 & ... & \alpha_i & ... & \alpha_{2n} \\ t_1 & ... & t_i & ... & t_{2n} \\ l_1 & ... & l_i & ... & l_{2n} \end{Bmatrix},$$

the proximity model is formalized as follows.

$$U = w_1 \cdot spatial\_proximity + w_2 \cdot temporal\_proximity \tag{7}$$

where $w_i$(i=1, 2) is a weight such that $w_i \in [0, 1)$ and $w_1 + w_2 = 1$. The *spatial_proximity* and *temporal_proximity* are defined as follows.

- *Spatial_proximity*: The *spatial_proximity* evaluates the average spatial proximity of all composition instances. We define *spatial_proximity* of a composition instance $S_k$ in Equation (8). Then, we average

*spatial_proximity* for the composition pattern $CP$ in Equation (9).

$$Spa = \sum_{i=1}^{n} \frac{1}{|x_i - x_{i+1}| + |y_i - y_{i+1}|} \quad (8)$$

$$spatial\_proximity = \frac{\sum_{j=1}^{sup(Seq)} Spa_j}{sup(Seq)} \quad (9)$$

where $n$ is the number of component IoT services corresponding to the composition instance, $l_i = <x_i, y_i>$ and $l_{i+1} = <x_{i+1}, y_{i+1}>$ are two coordinates for two consecutive component IoT services, and $sup(Seq)$ is the total number of composition instances. For instance, for the composition instance in Fig. 2(a), its *spatial_proximity* value is $Spa = \frac{1}{|l_2 - l_1|} = \frac{1}{|2-1|+|4-2|} = 0.33$. In this paper, Manhattan distance proposed in [24] is employed to evaluate the proximity because of its computing efficiency.

- *Temporal_proximity*: The *temporal_proximity* evaluates the average temporal proximity of all composition instances. We employ the technique proposed in [22] for computing the distance between time-interval based data. For the composition instance $S_k$, its component IoT service is $s_i = \{(s_i^+, st_i, sl_i), (s_i^-, et_i, el_i)\}$. We use a function $f_i$ with respect to $t$ to map the temporal information of $s_i$. $f_i$ is defined as follows.

$$f_i(t) = \begin{cases} 1, & t \in [st_i, et_i] \\ 0, & otherwise \end{cases} \quad (10)$$

Then we have a set of functions $\{f_1, f_2, ... f_n\}$ corresponding to the composition instance $S_k$. The *temporal_proximity* for the composition instance $S_k$ is computed by Equation (11). The average *temporal_proximity* for the composition pattern $CP$ is computed by Equation (12).

$$Temp = \frac{\int_{t_1}^{t_{2n}} \sum_{i=1}^{n} f_i(t)\, dt}{(t_{2n} - t_1) \cdot n} \quad (11)$$

$$temporal\_proximity = \frac{\sum_{j=1}^{sup(Seq)} Temp_j}{sup(Seq)} \quad (12)$$

where $t_1$ and $t_{2n}$ are the respective first and the last time point of $S_k$, and $n$ is the number of component IoT services in $S_k$. For instance, the *temporal_proximity* value for the composition instance { <stove, [18:00, 19:00]>, <washing machine, [18:40, 19:20 ] >} is computed by $\frac{(18:40-18:00)+(19:00-18:40)\cdot 2+(19:20-19:00)}{(19:20-18:00)\cdot 2} = 0.625$. This composition instance can be explained as *the occupant is cooking while he/she is also doing laundry*. Another composition instance is {<stove, [18:00, 19:00]>, <fan, [18:00, 19:00]>} and its *temporal_proximity* score is 1. The latter composition instance can be considered to be more temporally proximate than the former.

### 3.1.3 Probabilistic Composition Pattern.
For an activity, the occupant may perform it in different manners, resulting in variations in similar composition patterns. For instance, consider the "breakfast preparation" activity, the occupant may not perform this activity in exactly the same way each time. Some of the steps might be changed e.g., making coffee instead of tea or varied in order. Some IoT services may be not used some time e.g., kettle service is not used for making Cappuccino. The variations of the activity may result in different composition patterns for the same activity. Two similar composition patterns with slightly different involved IoT services may come from the same activity. These variations may result in too many composition patterns, which become an obstacle to analyze and understand the occupant's actual daily activities. Therefore, there is a need for a more compressed and concise representation to capture such variation. We propose the probabilistic composition pattern to capture the variation of the activity by introducing involvement probability.

After discovering composition patterns, we cluster them to get a more compact representation (i.e., probabilistic composition pattern). Specifically, we first group similar composition patterns into different clusters and then the probabilistic composition pattern corresponding to each cluster is generated. We employ the Jaccard similarity measure [4] to calculate the similarity between two composition patterns as follows.

$$similarity(CP_i, CP_j) = \frac{|Seq_i \cap Seq_j|}{|Seq_i \cup Seq_j|} \tag{13}$$

where $Seq_i$ and $Seq_j$ are the frequent sequences corresponding to the composition patterns $CP_i$ and $CP_j$. The clusters are formed to optimize an objective function in Equation (14) [42] so that the composition patterns within a cluster are similar to one another and dissimilar to ones in other clusters in terms of the frequent sequences.

$$F = \sum_{i=1}^{K} \sum_{CP_j \in cluster_i} (1 - similarity(CP_j, center_i)) \tag{14}$$

where $K$ is the number of clusters and $cluster_i$ is the $i$th cluster. The $center_i$ is the center for the $cluster_i$ and it is defined by the shared event set in the $cluster_i$.

**Definition 6: Probabilistic composition pattern.** Given $CP_i = \{S_i, Seq_i, sup(Seq_i)\}$ and $cluster_i = < CP_1, CP_2 ... CP_m >$ consisting of $m$ similar composition patterns, its corresponding probabilistic composition pattern is defined as $Pro\_CP = \{< \alpha_1, p_1 >, < \alpha_2, p_2 > ... < \alpha_n, p_n >\}$ where $< \alpha_i, p_i >$ denotes the probability that the occupant will use an IoT service and $\alpha_i$ is the corresponding service event and $p_i$ is the involvement probability formalized in Equation (15).

$$p_i = \frac{|\alpha_i|}{\sum_{i=1}^{m} sup(Seq_i)} \tag{15}$$

where $|\alpha_i|$ is the number of composition instances where $\alpha_i$ participates in and $\sum_{i=1}^{m} sup(Seq_i)$ is the total number of composition instances associated with $cluster_i$.

*3.1.4 Periodic probabilistic composition patterns.* The occupant generally performs some activities regularly. For instance, the occupant has the habit of taking a shower around 10pm in the bathroom. In this section, we introduce the novel notion of periodic probabilistic composition patterns to model the regularity of repeating composition patterns.

Definition 7: Periodic probabilistic composition patterns. A periodic probabilistic composition pattern *Per_Pro_CP* is defined as the repeating composition patterns at a particular location with regular time intervals. It can be described by the probabilistic composition pattern associated with time intervals and locations. It is denoted by a tuple *Per_Pro_CP* =< *Pro_CP,T,L,P* > where:

- $Pro\_CP$ is the probabilistic composition pattern defined in Definition 6.
- $T = < T_s, T_e >$ is a representative time interval associated with $Pro\_CP$, where $T_s$ and $T_e$ are the start time and end time, respectively. We collect all the time intervals for $Pro\_CP$ in $DB$, which is the set $\tau = \{< st_1, et_1 >, < st_2, et_2 > ... < st_m, et_m >\}$. The representative time interval $< T_s, T_e >$ can be calculated by minimizing its dissimilarity $dis$ between the time interval instance $< st_i, et_i >$. For example, the dissimilarity between the time intervals (3,5) and (4,6) as well as (7,8) are |3-4|+|5-6|=2 and |3-7|+|5-8|=7, respectively. Hence, the time interval (4,6) is more similar with (3,5) than that of (7,8). The dissimilarity $dis$ between two time intervals is defined as follows.

$$dis = |T_s - st_i| + |T_e - et_i| \tag{16}$$

Thus, the total dissimilarity between $< T_s, T_e >$ and $\tau$ is formalized as:

$$Dis(T, \tau) = \sum_{i=1}^{m} |T_s - st_i| + |T_e - et_i| \tag{17}$$

For the purpose of minimizing $Dis(T, \tau)$, Equation (17) is transformed into two known minimization problems, that is, find $T_s$ and $T_e$ to minimize $\sum_{i=1}^{m}|T_s - st_i|$ and $\sum_{i=1}^{m}|T_e - et_i|$, respectively. $T_s$ is the median of the start time set $\{st_1, st_2...st_m\}$ and $T_e$ is the median of the end time set $\{et_1, et_2...et_m\}$. The proof is provided in [25].

- $L$ is a location of $Pro\_CP$. We use regions such as the bedroom and the bathroom as the location.
- $P$ is the probability that $Pro\_CP$ occurs around time interval $T$ at location $L$. Suppose the time information of a $Pro\_CP$ instance (i.e., or a composition instance) is $< st_j, et_j >$. The $Pro\_CP$ is said to occur around time interval $T$ if their dissimilarity $dis$ is no more than a tolerance threshold $\zeta$, that is, $|T_s - st_j| + |T_e - et_j| \leq \zeta$. P is defined as follows.

$$P = \frac{Num}{TNum} \tag{18}$$

where $Num$ is the number of $Pro\_CP$ occurrence around time interval $T$ at location $L$. $TNum$ is the total number of $Pro\_CP$ occurrence in the database.

## 3.2 Temporal Relationships among Composition Patterns

It is important to understand the temporal relationships among composition patterns. The temporal relationships provide rich knowledge regarding how the occupants allocate time across activities. However, discovering temporal relationships among composition patterns is difficult. This is because the relationships among intervalbased composition patterns are much more complex than point-based data. Thus a new model that can capture the complex relationships among time intervals is required. Allen's temporal logic is a commonly used model to describe the complex temporal relationships among time intervals. In this paper, we adapt Allen's temporal logic to model the temporal relationships among composition patterns.

Definition 8: Temporal relationships among composition patterns. Given two composition patterns $CP_i$ and $CP_j$, $< S_i, [st_i, et_i] >$ and $< S_j, [st_j, et_j] >$ are the composition instances associated with time intervals corresponding to $CP_i$ and $CP_j$, respectively. The temporal relationships among composition patterns can be represented by a $n \times n$ matrix $M$ whose element $R[i, j] =< tran\_pro, r >$ represents temporal relationships between any two pairwise composition patterns where

- $r$ denotes the temporal relationships between two composition patterns, which is formalized in Fig.4.
- $tran\_pro$ denotes the probability of composition pattern $CP_i$ transiting to $CP_j$. $CP_i$ is said to transit to $CP_j$ if there exist two corresponding composition instances $S_i$ and $S_j$ such that $st_i \leq st_j$. $tran\_pro$ is formalized as follows.

$$tran\_pro = \frac{Num_{ij}}{Num_i} \tag{19}$$

where $Num_{ij}$ is the number of composition instances in $CP_i$ transiting to the composition instances in $CP_j$ and $Num_i$ is the total number of composition instances in $CP_i$.

## 4 PREDICTION MODEL

The typical prediction problem basically focuses on training a model from labeled data. The trained model is used to estimate the label for new data. Instead of training a model from labeled data, we first discover knowledge from unlabeled data and use the knowledge to build the prediction model. Moreover, unlike the typical prediction problem which focuses on estimating the label for new data, our prediction model can not only forecast what activity (i.e., the label) will happen but also estimate when and where the activity will take place.

In this section, we aim at building the prediction component of Cognitive Amplifier for forecasting what, when, and where the occupant will do next. Specifically, we combine the discovered knowledge including periodic probabilistic composition patterns and temporal relationships among the composition patterns to build the prediction model $\lambda$. As shown in Fig. 5, the ovals represent periodic probabilistic composition patterns and the rectangles represent involved IoT service events. $b_{kl}$ represents the involvement probability. ($a_{ij}, r_{ij}$) represents transition probability between composition patterns and their temporal relationships, respectively. Therefore, the prediction problem can be formulated as: 'given the prediction model $\lambda$ and a new IoT service event sequence $O = o_1 o_2 ... o_t$, find the next most probable activity $q$ and its occurrence time interval $t$ as well as location $l'$, which is formalized as follows.

| Temporal relations | Formalization | Pictorial example |
|---|---|---|
| $S_i$ before $S_j$ | $et_i < st_j$ | 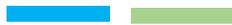 |
| $S_i$ overlap $S_j$ | $st_i < st_j < et_i < et_j$ | 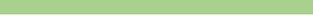 |
| $S_i$ equal $S_j$ | $st_i = st_j$ and $et_i = et_j$ | 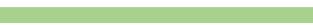 |
| $S_i$ start-by $S_j$ | $st_i = st_j$ and $et_i < et_j$ | 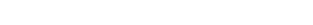 |
| $S_i$ finish $S_j$ | $st_i > st_j$ and $et_i = et_j$ | 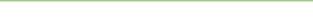 |
| $S_i$ meet $S_j$ | $et_i = st_j$ | 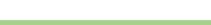 |
| $S_i$ during $S_j$ | $st_j < st_i < et_i < et_j$ | 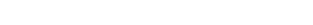 |

Fig. 4. Temporal relationships among two composition instances

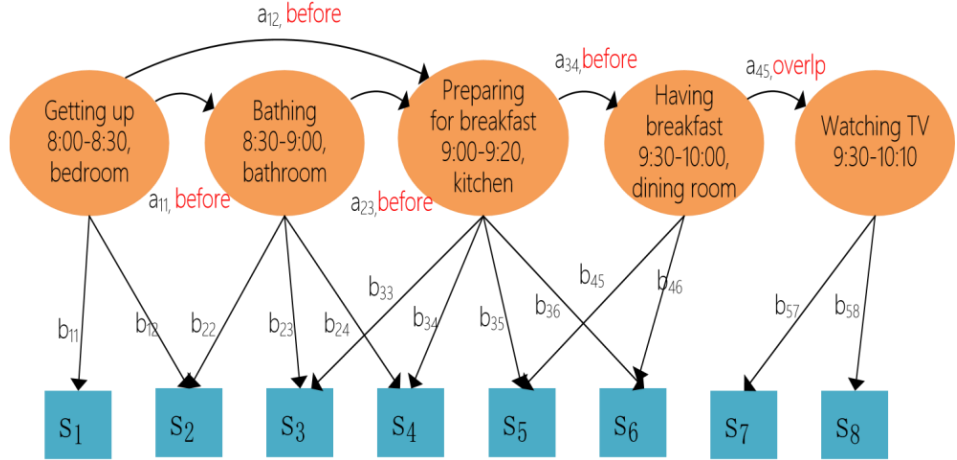

Fig. 5. The predictive model.

$$\delta(q,t,l) = \max P(q,t,l|\lambda,o_1 o_2...o_t) \quad (20) \quad q,t,l$$

The prediction problem can be divided into two subproblems. We first recognize the label of the ongoing activity $O$ based on the periodic probabilistic composition patterns. Then, the next activity $(q,t,l)$ is inferred based on the temporal relationships among composition patterns.

In the first subproblem, we recognize the label of the ongoing activity $O$ through comparing the similarity between $O$ and the periodic probabilistic composition pattern $i$ as follows.

$$Y(i,O) = Y_S(i,O) + Y_T(i,O) + Y_L(i,O) \tag{21}$$

In Equation (21), $Y_S(i,O)$ refers to the structure similarity (i.e., $i$ and $O$ are similar in terms of involved IoT services.). We use the Euclidean distance to measure the structure similarity. $Y_T(i,O)$ refers to time similarity ( i.e., $i$ and $O$ are similar if they happen at similar times). We use Equation (11) to compute the time similarity. $Y_L(i,O)$ refers to the location similarity (i.e., $i$ and $O$ are similar if they happen in similar locations. ). For simplicity, we set the similarity values to have equal effects; however, it is possible to define $Y(i,O)$ as a weighted average. In the second subproblem, we infer the next activity $(q,t,l)$ based on the predicted outcome $i$ in the first subproblem. According to the prediction model as shown in Figure 5, the next activity $(q,t,l)$ can be easily inferred based on the temporal relationships among composition patterns.

## 5   APPROACH

There are three subtasks for implementing the Cognitive Amplifier including the discovery of periodic probabilistic composition patterns and temporal relationships among composition patterns as well as the prediction task. In the first subtask, we focus on finding all the periodic probabilistic composition patterns. Then, we find temporal relationships among the composition patterns. Lastly, we develop a prediction algorithm based on the outcomes of stage one and two to infer the label according to the approach in section 4. The first and second subtasks are detailed as follows.

In the first subtask, the algorithm *Per_Pro_CPMiner* is developed to search periodic probabilistic composition patterns from IoT service event sequences. The algorithm constitutes six stages. The algorithm starts with *dividing the search space* into multiple smaller searching space. Second, *Per_Pro_CPMiner finds all composition patterns* in a smaller searching space. Next, *Per_Pro_CPMiner* removes low-quality composition patterns using the *significance* and *proximity* measures. For the purpose of clarity, these three steps are implemented by an algorithm *CPMiner*, which is the key sub-procedure of the *Per_Pro_CPMiner*. The *CPMiner* algorithm focuses on discovering proximate and significant composition patterns. Fourth, the remaining composition patterns are clustered into multiple groups. Fifth, for each group, *Per_Pro_CPMiner* calculates the involvement probability of

| | | | | | | |
|---|---|---|---|---|---|---|
| L₁ | (A·A):3 | (B·B):3 | (C·C):3 | (E·E):2 | (F·F):3 | |
| Projected database | (B·C·D·C·B·D·E·F·F·E·)<br>(B·M·C·C·B·K·M·K·E·F·F·E·)<br>(B·C·C·B·G·G·F·F·) | (C·D·C·D·E·F·F·E·)<br>(M·C·C·K·M·K·E·F·F·E·)<br>(C·C·G·G·F·F·) | (D·B·D·E·F·F·E·)<br>(B·K·M·K·E·F·F·E·)<br>(B·G·G·F·F·) | (F·F)<br>(F·F)<br>(F·F) | (E·)<br>(E·) | |
| Frequent event pairs | (B·B):3 (C·C):3<br>(E·E):2 (F·F):3 | (C·C):3 (E·E):2<br>(F·F):3 | (E·E):2 (F·F):3 | (F·F):3 | 0 | |
| L₂ | (A·B·B):3 | (A·C·C):3 | (A·E·E):2 | (A·F·F):3 | | |
| Projected database | (C·D·C·D·E·F·F·E·)<br>(M·C·C·K·M·K·E·F·F·E·)<br>(C·C·G·G·F·F·) | (D·B·D·E·F·F·E·)<br>(B·K·M·K·E·F·F·E·)<br>(B·G·G·F·F·) | (F·F)<br>(F·F) | Null<br>Null | | |
| Frequent event pairs | (C·C):3 (E·E):2<br>(F·F):3 | (E·E):2 (F·F):3 | (F·F):2 | 0 | | |
| L₃ | (A·A·B·C·C·B):3 | (A·A·B·B·E·E):2 | (A·A·B·B·F·F):3 | (A·A·C·C·E·E):2 | (A·A·C·C·F·F):3 | (A·A·E·F·F·E):2 |
| Projected database | (D·E·F·F·E·)<br>(K·M·K·E·F·F·E·)<br>(G·G·F·F·) | (F·F)<br>(F·F) | (E·)<br>(E·) | (F·F)<br>(F·F) | (E·)<br>(E·) | Null |
| Frequent event pairs | (E·E):2 (F·F):3 | (F·F):2 | 0 | (F·F):2 | 0 | 0 |
| L₄ | (A·A·B·C·C·B·E·E):2 | (A·A·B·C·C·B·F·F):3 | (A·A·B·E·F·F·E):2 | (A·A·C·C·E·F·F·E):2 | | |
| Projected database | (F·F)<br>(F·F) | (E·)<br>(E·) | Null | Null | | |
| Frequent event pairs | (F·F):2 | 0 | 0 | 0 | | |
| L₅ | (A·A·B·C·C·B·E·F·F·E):2 | | | | | |
| Projected database | Null | | | | | |
| Frequent event pairs | 0 | | | | | |

Fig. 6. The process of *CPMiner*

each event and generate *probabilistic composition patterns*. Finally, for each probabilistic composition patterns generated in the fifth phase, the algorithm finds its associated time and location information. This information is used to estimate the representative time intervals and locations corresponding to each probabilistic composition pattern. A set of *periodic probabilistic composition patterns* is generated in this phase. For the purpose of coherence with the terminology from data mining techniques, *event patterns* is used to refer to composition patterns. The example in Fig. 2 to used to explain the process of *CPMiner* as illustrated in Fig.6

*Phase I: Dividing search space*. A smart home usually has one or more regions such as a bathroom and a living room. An IoT service is located in a particular region. Thus, its generated event sequence is associated with the region. For instance, a "turn on the TV" event occurs in the living room. For the smart home's layout, it consists of a region set $r = \{r_1, r_2...r_n\}$. The *DB* containing IoT service event sequence database is divided into multiple smaller databases $DB_{r_i}$. The $DB_{r_i}$ contains events that occur only in the region $r_i$. In the remaining phases, the searching process for composition patterns is performed on each smaller databases. For the sake of explaining the process of *CPMiner*, we assume that all events shown in Fig. 2 are collected in the same region.

*Phase II: Searching event patterns*. *CPMiner* adopts a divide-and-conquer, pattern-growth principle from Prefixspan [26]. The principle is explained as follows: given an event pattern, the databases are recursively projected into multiple smaller *projected databases*. Then the given event pattern is extended through searching the smaller projected databases instead of searching the whole database.

*Definition 8: Projected database*. Let $p$ be an event pattern in a database *DB*. The $p$-projected database, denoted as $DB|_p$, is the set of suffixes of event sequences in *DB* with regard to the prefix $p$.

The searching process for event patterns constitutes three steps. In the first step, the algorithm searches a collection of event patterns with one length, resulting in a 1-length event patterns set. Next, for each 1-length event pattern, associated projected databases are constructed. Then, the length of each event pattern is grown through searching its associated projected databases. The three steps are detailed as follows.

1.    *Search the set of 1-length event patterns $L_1$*. For a database such as the one in Fig. 2, *CPMiner* starts with counting the frequency of each event pairs and removes those with low frequency. For instance, if we set the *minsup* threshold to be 2, the 1-length event patterns whose frequency is more than 2 remain and consist of the 1-length event pattern set $L_1$. For instance, $(A^+A^-)$: 3 represents the event pattern and its frequency as shown in Fig. 6.

2.    *Construct projected databases for each 1-length event pattern*. Suppose $L_1 = \{\alpha_1^1, \alpha_2^1...\alpha_n^1\}$ is the set 1-length event patterns set. For an element $\alpha_i^1$, an associated projected database $DB|_{\alpha_{i1}}$ is constructed. According to the definition 7, $DB|_{\alpha_{i1}}$ is the set of suffix event sequences based on its prefix $\alpha_i^1$.

3.    *A k-length event pattern $\alpha$ is extended to the (k+1)-length event pattern $\alpha^{'}$ by searching its projected database $DB|_\alpha$   ( $k \geq 1$ )*. Given a prefix $\alpha$ event pattern, *CPMiner* finds the local frequent event pairs through searching its projected database $DB|_\alpha$ . The local frequent event pairs constitutes the set $\{e_1, e_2...e_n\}$ and infrequent ones are removed. Then, the frequent event pair $e_i$ is appended to the prefix $\alpha$, resulting a new frequent event pattern $\alpha^{'}$ with the length increased by 1. Thus, the (k+1)-length event patterns prefixed with $\alpha$ are generated. Not that we consider event pairs and single events in the projected database will not be considered again.

An example of finding event patterns prefixed with $(A^+A^-)$ is explained as follows. The projected database for $(A^+A^-)$ is $DB|_{(A+A-)}$. We can find local frequent event pairs by scanning $DB|_{(A+A-)}$. They are $(B^+B^-$: 3), $(C^+C^-$: 3), $(E^+E^-$: 2), and $(F^+F^-$: 3). By appending the local frequent event pairs to the prefix $(A^+A^-)$, we have 2-length event patterns $L_2$. They are $(A^+A^-B^+B^-$: 3), $(A^+A^-C^+C^-$: 3), $(A^+A^-E^+E^-$: 2), and $(A^+A^-F^+F^-$: 3). For each 2-length event patter in $L_2$, we perform the projection and appendage recursively, we can find 3-length event patterns. For example, for $(A^+A^-B^+B^-)$, we can find its local frequent event pairs by searching its projected database. These local frequent event pairs are $(C^+C^-$: 3), $(E^+E^-$: 2), and $(F^+F^-$: 3). We append $(C^+C^-$: 3), $(E^+E^-$: 2), and $(F^+F^-$: 3) to the prefix $(A^+A^-B^+B^-)$. Thus, we can get the 3-length event patterns including $(A^+A^-B^+C^+C^-B^-$: 3), $(A^+A^-B^+B^-E^+E^-$: 2), and $(A^+A^-B^+B^-F^+F^-$: 3). By the same method, we have $(A^+A^-C^+C^-E^+E^-$: 2) and $(A^+A^-C^+C^-F^+F^-$: 3) prefixed with $(A^+A^-C^+C^-)$, and $(A^+A^-E^+F^+F^-E^-$: 2) prefixed with $(A^+A^-E^+E^-)$. Similarly, we can find $L_4$ and $L_5$ by the same method.

Algorithm 1: *Per_Pro_CPMiner* algorithm

Input: A spatio-temporal database ($DB$), region set $r = \{r_1, r_2...r_n\}$, support threshold ($minsup$), significance threshold ($minsi\vartheta$), proximity threshold ($minpro$), weight of the *spatial_proximity* ($w_1$), and weight of the *temporal_proximity* ($w_2$).

Output: A set of periodic probabilistic event patterns.

Procedure Divide_Search_Space ($DB, r$)

1: transform $DB$ into $|r|$ numbers of sub-databases based on $r$;

2: return $DB_{r_i}$;

Procedure Search_Event_Pattern ($DB^r_i$, $minsup$)

3: $L_1$ = find_1-length_pattern($DB^r_i$, $minsup$ );

4: Prefixspan($\alpha, l, DB^r_i|_\alpha$)//$\alpha$ is an event pattern, $l$ is the length of $\alpha$ ($l \geq 1$), $DB^r_i|_\alpha$ is the $\alpha$-projected database.

5: Scan $DB^r_i|_\alpha$ once, find the set of local frequent events $\{e_1, e_2...e_n\}$;

6: for each frequent event $e_l$ do

7: $\alpha' = \alpha + e_l$; //append $e_l$ to $\alpha$ to form the new event pattern $\alpha'$.

8: for each $\alpha'$ do

9: add $\alpha'$ to the event pattern set *patternset*;

10: construct $\alpha'$-projected database $DB^r_i|_{\alpha'}$, and call Prefixspan ($\alpha', l + 1, DB^r_i|_{\alpha'}$);

11: return *patternset*;

Procedure Compute_Significance_Proximity (*patternset*, $minsi\vartheta$, $minpro$, $w_1$, $w_2$);

12: for each pattern $\alpha$ in *patternset* do

13: computeSignificance($\alpha$, $minsi\vartheta$); // Calculate significance for each pattern by Definition 4 and discard those patterns whose significance are less than $minsi\vartheta$.

14: computeProximity ( $\alpha$, $w_1$, $w_2$, $minpro$);// Calculate proximity for each pattern by Definition 5 and discard those patterns whose proximity are less than $minpro$.

15: return *patternset'* ;

Procedure Cluster_Event_Patterns (*patternset'*, $K$);

16: arbitrarily choose $K$ event patterns as the initial cluster centers;

17: repeat

18: assign each event pattern to the cluster to which the event pattern is most similar with the center;

19: update the cluster center, that is, calculate the shared events of event patterns for each cluster;

20: until center does not change;

21: return a set of clusters *clusters* = {*cluster$_1$*, *cluster$_2$*...*cluster$_k$*};

Procedure Generate_Probabilistic_Event_Patterns (*clusters*);

22: for each *cluster$_i$* in *clusters* do

23: collect all event types and compute their corresponding involvement probability by Equation (15);

24: return a set of probabilistic event patterns *Pro_CP*;

Procedure Generate_Periodic_Probabilistic_Event_Patterns (*Pro_CP*, $DB^r_i$);

25: for each pattern $\alpha$ in *Pro_CP* do

26: find associated time and location information from $DB^r_i$;

27: estimate the time interval $T$, location $L$, and probability $P$ for $\alpha$ by Definition 7;

28: add $< \alpha, T, L, P >$ to the set *Per_Pro_CP*;

29: return a set of periodic probabilistic event patterns *Per_Pro_CP* ;

*Phase III: Calculate significance and proximity for event patterns*. Many of the discovered event patterns in phase II may be of low-quality. We employ the quality measure in terms of significance and proximity to prune those low quality event patterns. The significance of each event pattern is computed by Definition 4. We discard the event patterns whose significance scores are less than the specified significance threshold $minsi\partial$. In addition, we compute the proximity score for a given event pattern by Definition 5. We discard the event patterns whose proximity scores are less than the specified proximity threshold $minpro$. For instance, we set the weight $w_1$ and $w_2$ for *spatial_proximity* and *temporal_proximity* to be 0 and 1, respectively. We can have the proximity scores for the 2-length event patterns, they are, $<A^+A^-B^+B^->$:0.418, $<A^+A^-C^+C^->$:0.358, $<A^+A^-E^+E^->$:0.098, $<A^+A^-F^+F^->$:0.082, $<B^+C^+C^-B^->$:0.906, $<B^+B^-E^+E^->$:0.224, $<B^+B^-F^+F^->$:0.215, $<C^+C^-E^+E^->$:0.209, $<C^+C^-F^+F^->$:0.199, $<E^+F^+F^-E^->$:0.879. We specify the proximity threshold $minpro$ to be 0.3. Thus, only the event patterns whose proximity scores are more than 0.3 remain. The final results of our running example are $<E^+F^+F^-E^->$ and $<A^+A^-B^+C^+C^-B^->$. Their proximity scores are 0.879 and 0.494, respectively.

*Phase IV: Cluster event patterns into multiple groups*. After applying the quality measures in phase III, we have high- quality event patterns. Some of these event patterns may be similar because they are the variations of the same activity. As a result, a clustering technique is applied to group these similar event patterns. We borrow the idea from *Kmeans* clustering algorithm to find similar event patterns. The *Kmeans* algorithm is efficient in clustering numeric data points but could not be directly applied to cluster event sequences. The traditional *Kmeans* algorithm employs Euclidean distance to compute the similarity between data points and updates the center by computing the mean within the cluster [42]. In this paper, we use Jaccard similarity measure to compute the similarity between event patterns and use shared events as the center.

Given the set of event patterns and the number of clusters *K*, our goal is to is to find an assignment set $cluster_i$ for each event pattern, and the center $center_i$ of each cluster to minimize the objective function *F* in Equation (14). We start with the *Kmeans* algorithm by choosing *K* event patterns as *K* centers randomly. The *Kmeans* algorithm proceeds by alternating between two steps iteratively: assignment step and update step. In the assignment step, an event pattern is assigned to the cluster with the closest center based on the similarity in Equation (13). After finding the new assignment set $cluster_i$ for each event pattern, we calculate the new center for each $cluster_i$ in the update step, according to the shared events among all event patterns in $cluster_i$. After updating many times, the algorithm will converge when the assignment no longer changes.

*Phase V: Generating probabilistic event patterns*. After performing the clustering step, we can obtain *K* clusters {$cluster_1, cluster_2...cluster_k$ }. For each cluster $cluster_i$, a corresponding probabilistic composition pattern is generated. We collect all event types in the cluster$cluster_i$ and compute the corresponding involvement probability for each event type by Equation (15). Thus we can get the probabilistic composition patterns.

*Phase VI: Generating periodic probabilistic event patterns*. In this phase, the algorithm focuses on computing the representative time intervals and a location for each probabilistic pattern by Definition 7. The final result of phase VI is a set of probabilistic event patterns associated with time intervals, location, and a probability (i.e., periodic probabilistic event patterns).

In the second subtask, we propose the TPminer algorithm to discover temporal relationships among composition patterns. This stage consists of two subtasks. The first subtask focuses on computing the transition probability between two composition patterns. For the sake of clarity, all the supporting composition instances for a composition pattern is replaced by a symbol. For example, the composition instance $<A^+A^-B^+B^->$ is replaced by $S_1$. Therefore, we can easily compute the transition probabilities among composition patterns by Equation (19). After getting the transition probabilities among composition patterns, the second subtask focuses on detecting their temporal relationships based on temporal logic defined in Fig. 4. The *Tem_Rel_CPMiner* is detailed in Algorithm 2.

---

Algorithm 2: *Tem_Rel_CPMiner* algorithm

---

Input: An IoT service event sequence database (*DB*), a set of event patterns (*patternset'*). Output: A temporal matrix *M*.

1: for each $\alpha$ in *patternset'* do
2: find all supporting composition instances for $\alpha$ in *DB* and replace these instances with the symbol $S_i$; 3: for $S_i$ and $S_j$ do
4: *tran_pro* = computeTransitionProbability($S_i$, $S_j$);//Calculate the transition probability between $S_i$ and $S_j$ by Equation (19).
5: *r* = computeTemporalRelationship($S_i$, $S_j$);// Calculate the temporal relationships between $S_i$ and $S_j$ based on temporal relationships defined in Fig.4. 6: return A temporal matrix *M*.

---

## 6   EXPERIMENTAL RESULTS

We conduct a set of experiments on a 3.4 GHZ Intel processor and 8 GB RAM under Windows 10 environment to systematically evaluate our proposed approach. The experiments are performed on four real datasets (i.e., Data1, Data2, Data3, and Data4). Data1 and Data2 and Data3 are collected in a smart home environment setting from CASAS project [44]. These three datasets are organized in the same format < Date, Time stamp, Sensor ID, ON/OFF> (e.g., < 2008-02-21, 11:45:35, M14, ON > ). For the location, Data1 and Data2 and Data3 are collected in different locations. Data4 collects an occupant's natural interaction with daily things without any interruption. It records things usage for a single occupant for two weeks [43]. The data format of Data4 is <ID, Start time, End time, Location> (e.g., <Light, 8:00, 9:00, Bedroom>). Moreover, all activities in the four datasets are annotated with labels. Specifically, Data1, Data2, Data3 and Data4 have 5, 8, 15, and 23 activities, respectively.

We conduct five sets of experiments. The first experiment focuses on evaluating the performance and scalability of *Per_Pro_CPMiner* in discovering composition patterns. The second experiment aims to evaluate the effectiveness of our quality criteria (i.e., significance and proximity) in pruning low-quality composition patterns. The third experiment focuses on evaluating the applicability of the *Per_Pro_CPMiner* and *Tem_Rel_CPMiner* by showing some results of discovered periodic probabilistic composition patterns and temporal relationships among these patterns, respectively. The fourth set is to evaluate prediction accuracy. The last experiment is to evaluate how much convenience can be obtained by implementing the proposed approach.

We conduct the first set of experiments on a combined four datasets. The support threshold *sup* is set to be varied from 5% to 10%. Fig. 7(a) illustrates the execution time in discovering composition patterns. The execution time decreases with an increase in the support threshold. Fig. 7(b) shows that the number of composition patterns decreases with an increase in the support. These are expected results because the increasing support enables the algorithm to narrow down the scope in each iteration through discarding the low frequent composition patterns.

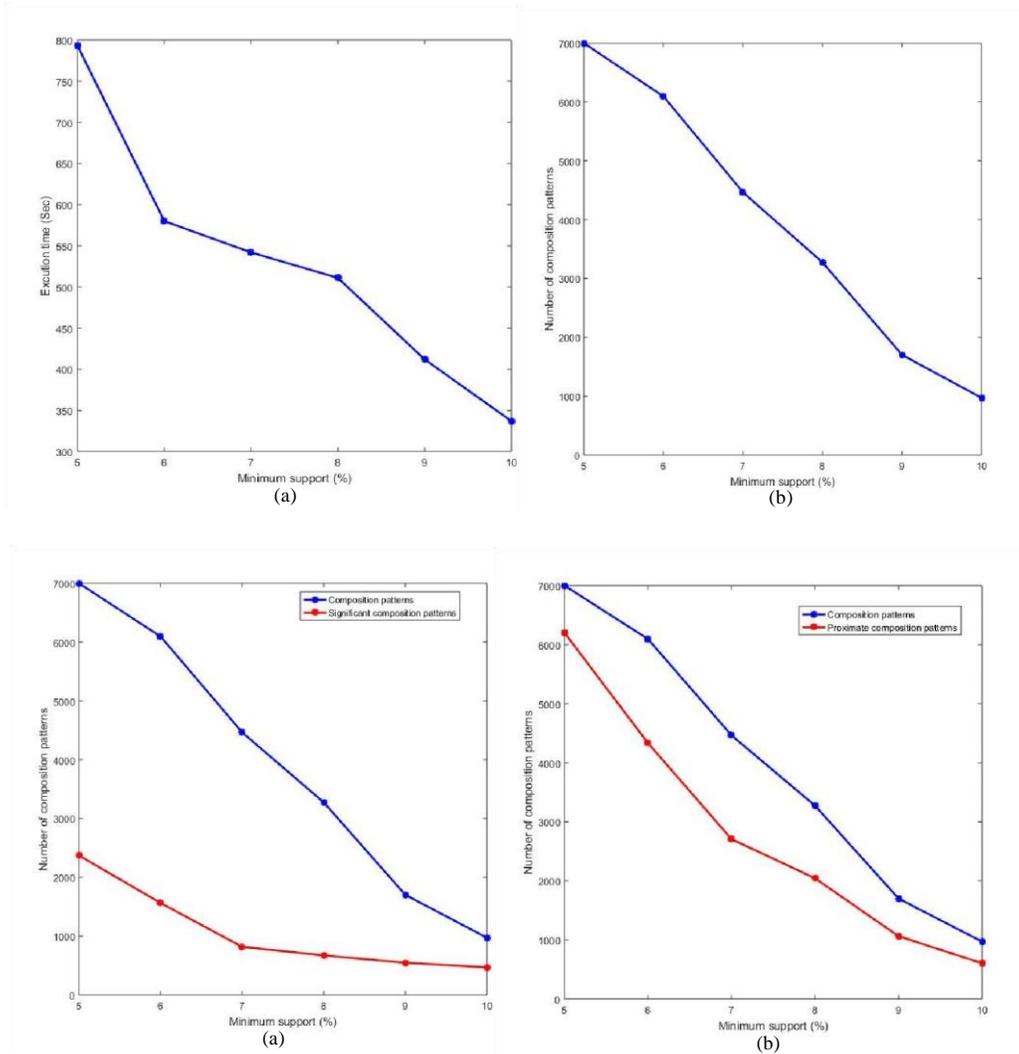

Fig. 8. Effectiveness of significance and proximity

| Labels | Involvement probability | | Periodic features |
|---|---|---|---|
| Taking medication | (refrigerator,1) | (light, 0.42) | (1:54-2:00, in the kitchen, 0.5) (19:23-19:38, in the kitchen, 0.29) |
| Preparing breakfast | (refrigerator, 0.89) | (garbage disposal, 0.72) | (5:45-6:20, in the kitchen, 0.75) |
| | (light,0.83) | (microwave,0.78) | |
| | (door, 0.78) | (cabinet, 0.56) | |
| | (sink faucet, 0.39) | (toaster, 0.28) | |
| Preparing lunch | (microwave, 1) | (refrigerator, 0.85) | (11:15-12:05, in the kitchen, 0.64) |
| | (light, 0.65) | (sink faucet, 0.65) | |
| | (toaster, 0.6) | (cabinet, 0.6) | |
| | (burner, 0.2) | (garbage disposal,0.55) | |
| Preparing dinner | (refrigerator, 0.93) | (light, 0.86) | |

| | (door, 0.64) | (cabinet, 0.64) | (18:15-18:48, in the kitchen, 0.5) |
|---|---|---|---|
| | (microwave, 0.57) | (sink faucet, 0.5) | |
| | (toaster, 0.29) | (burner, 0.29) | |
| Watching TV | (TV,1) | (light,0.6) | (7:00-7:23, in the living room, 0.27) |
| | (sink,0.47) | (microwave,0.33) | (14:27-15:22, in the living room, 0.33) |
| | (refrigerator, 0.27) | (toilet, 0.2) | |
| Preparing snacks | refrigerator: 0.69 | light: 0.44 | |
| | toaster: 0.38 | sink: 0.38 | (8:49-9:29, in the kitchen, 0.31) |
| | cabinet: 0.38 | microwave: 0.31 | (20:02-21:27, in the kitchen, 0.25) |
| Listening to music | stereo: 1 | refrigerator: 0.56 | |
| | light: 0.56 | microwave: 0.5 | (9:19-11:00, in the living room, 0.5) |
| | sink: 0.38 | toaster: 0.25 | |
| Toileting | toilet: 1 flush: 1 | garbage disposal: 0.4 | Null |
| Washing dishes | sink: 1 | cabinet: 0.45 | |
| | refrigerator: 0.3 | microwave: 0.25 | (7:08-7:34, in the kitchen, 0.45) |
| | light: 0.2 | garbage disposal: 0.25 | (11:28-11:31, in the kitchen, 0.35) |

Fig. 9. Primary discovered periodic probabilistic composition patterns

We conduct the second set of experiments to evaluate the effectiveness of the pruning strategies including *si∂nif icance* and *proximity* in discarding low-quality composition patterns. In this experiment, we also use the combined four datasets. The *si∂nif icance* threshold is set to be 0.01. We evaluate the effectiveness of *si∂nif icance* in pruning insignificant composition patterns. We vary the support threshold from 5% to 10%. Fig. 8(a) shows the total number of composition patterns discovered from datasets and the number of significant composition patterns after applying the *si∂nif icance* strategy. From the results, we can see that the *si∂nif icance* is effective in filtering out insignificant composition patterns. For instance, the number of composition patterns is reduced from 972 to 468 at the 10% support threshold by applying the *si∂nif icance* strategy. Next, we evaluate the effectiveness of *proximity* strategy in pruning loosely correlated composition patterns. We vary the support threshold from 5% to 10%. The weight of *spatial_proximity* is to be 0 because the coordinates of IoT services are not provided in the datasets. Thus, the weight of *temporal_proximity* is set to be 1. We set the *proximity* threshold as 0.39. Fig. 8(b) shows the total number of composition patterns discovered from datasets and proximate composition patterns after applying the *proximity* strategy. It is clear that the *proximity* strategy is effective in pruning loosely correlated composition patterns. For instance, when the support threshold is set to be 10%, the *proximity* strategy is able to reduce the number of composition patterns from 972 to 606. In a word, Fig. 8 shows the *si∂nif icance* and *proximity* strategies are promising quality measure in pruning low-quality composition patterns. This is an expected result because the two strategies enable the algorithm to discard non-promising composition patterns in each iteration and the search scope is shrunk for the next iteration.

We conduct the third set of experiments to evaluate the applicability of the proposed approach by showcasing some interesting results we have found in Data4. We cluster the discovered composition patterns into multiple clusters by *Kmeans*. The cluster number *K* of 9 is found to be a suitable value based on several runs of the *Kmeans*. Fig. 9 shows the primary discovered periodic probabilistic composition patterns.

Some composition patterns such as "lawn work" and "going out for entertainment" are difficult to be discovered because of their low occurrence frequency. For instance, these two composition patterns occur only once during two weeks. Next, we discuss the periodic features of composition patterns. We plan to associate one representative time interval with a composition pattern. However, we find that it is better to associate multiple representative time intervals. For instance, the "preparing breakfast" frequently occurs in the morning. Thus, it is natural to associate only one representative time interval. For the "taking medication" composition pattern, it occurs in the morning and in the evening. It's better to associate two representative time intervals to describe its periodical features. In this experiment, we devise a simple technique to handle such an issue. Specifically, we set the tolerance threshold $\zeta$ to be 2 hours and $P$ to be 0.25. Given a composition pattern, we divide its supporting composition instances into groups by computing the distance between time intervals. Two time intervals are grouped together if they overlap. As shown in Fig. 9, we can find that some composition patterns indeed have periodic features. For instance, a striking periodic composition pattern is the " preparing lunch". There is a 64% likelihood that the occupant will perform the " preparing lunch " activity during 11:15 and 12:05. For the " toileting" activity, there is no striking periodic feature found under the setting although it is the most frequent occurring activities. This is because the " toileting" activity is distributed more uniformly than other activities such as " taking medication". Finally, we discuss the variation feature of the activities. Indeed, we find that the occupant performs the same activity in different manners. For example, for the " watching TV", there is a 60% chance he/she will use the light while watching TV. We can also infer that the occupant may have the habit of eating while watching TV because he/she uses the sink, microwave, and refrigerator. Thus, we can conclude that the involvement probability is a powerful concept which is not only able to capture the variations of activities but also describe interleaved activities.

We also showcase some temporal relationships among composition patterns found in Data4 as shown in Fig. 10. We can see that the occupant performs most of his/her daily activities in a sequential manner. For example, the occupant sometimes performs the" preparing breakfast" before " preparing snacks". However, the occupant has the habit of performing entertaining activities (i.e., " watching TV" and " listening to music") with other activities in a parallel manner. For example, he/she sometimes prepares breakfast while watching TV.

The fourth set of experiment is to evaluate the prediction accuracy for the prediction model. The prediction model is constructed based on the knowledge including periodic probabilistic composition patterns and the temporal relationships among the composition patterns which is found in the third experiment. The prediction process consists of two parts. The first part focuses on recognizing the ongoing activity based on the periodic probabilistic composition patterns given a new IoT service event sequence. The second part is to predict the next activity based on the temporal relationships among composition patterns given the recognized activity in the first part. For the first part, Fig. 11 shows the prediction accuracy for the primary discovered activities. It should be noted that the periodic feature is useful in differentiating the activities whose involved IoT services are similar. For example, the " preparing breakfast", " preparing lunch", and " preparing dinner" are quite similar in terms of their involved IoT services. Actually, the prediction accuracy for the three activities is respectively 72%, 54%, 80% without considering periodic features. After considering the periodic features, the prediction accuracy for these three activities is 90%, 81%, and 92%, respectively as shown in Fig. 11. However, the periodic feature has negative effects on recognizing the " toileting" because this activity does not have a striking periodic feature. In addition, some similar activities might be mistaken with one another, such as " preparing snacks" and " preparing breakfast" which happen at similar times and similar locations as well as similar involved IoT services. That's why the prediction accuracy for the " preparing snacks" is only 43%. However, one can still observe that despite the fact that the real-life activities are performed in various manners, our approach can still recognize a considerable number of activities. For the second part, one can easily predict the next activity based on the temporal relationships as shown in Fig.10. For example, if the recognized activity is " preparing breakfast", the next probable activity is "washing dishes". One can further predict that the "washing dishes" will happen around 7:08am-7:34am in the kitchen with 0.45 confidence as shown in Fig.9. Therefore, we can conclude that our approach is able to predict what, when, and where the next activity will happen.

| | preparing breakfast | preparing snacks | preparing lunch | preparing dinner | listening to music | watching TV | toileting | taking medication | washing dishes |
|---|---|---|---|---|---|---|---|---|---|
| preparing breakfast | | before:0.18 | | | before:0.06 | before:0.125 overlap:0.06 | before:0.25 | | before:0.325 |
| preparing snacks | | | before:0.08 | before:0.17 | during:0.08 before:0.25 overlap:0.08 | overlap:0.08 | before:0.08 | before:0.08 | before:0.08 |
| preparing lunch | | | | | before:0.25 | before:0.125 | before:0.125 | | before:0.5 |
| preparing dinner | | before:0.25 | | | | | | before:0.5 | before:0.25 |
| listening to music | during:0.08 | before:0.17 | before:0.25 overlap:0.08 | before:0.25 | | | | | before:0.17 |
| watching TV | | before:0.42 | before:0.17 | before:0.17 | before:0.08 | | before:0.08 | | before:0.08 |
| toileting | before:0.19 | before:0.09 | before:0.14 | before:0.05 | during:0.05 before:0.09 overlap:0.05 | before:0.09 during:0.09 | | before:0.09 | before:0.05 |
| taking medication | before:0.42 | before:0.17 | | before:0.08 | | before:0.08 | before:0.25 | | |
| washing dishes | | before:0.07 | before:0.27 | | during:0.07 before:0.2 | during:0.07 overlap:0.07 | before:0.25 | | |

Fig. 10. Primary discovered temporal relationships among composition patterns

In the last experiment, we evaluate how much convenience can be achieved by implementing the proposed approach. This experiment is based on the results discovered in the fourth experiment. We measure the convenience in terms of saving efforts and time. We assume some IoT services are remotely accessible such as the TV, the air conditioning, and the blind. We synthesize waiting time for some IoT services ranging from 1-3 minutes. We randomly choose five days in Data4 to test how much convenience can be obtained. The saved convenience in terms of saved efforts and time is shown in Fig.12. For example, the average saved efforts and time for Day1 is 8.4% and 35 minutes, respectively. The saved efforts highly depends on the number of IoT services that are assumed to be remotely controllable. Similarly, the saved time for each day depends on the number of IoT services that are assumed to have waiting time and the length of waiting time.

## 7  RELATED WORK

### 7.1  Service mining

Service mining is defined as the process of mining interesting and meaningful relationships among services [27]. An efficient algorithm *CoPMiner* is proposed to discover meaningful temporal relations among appliances [28]. The algorithm *CoPMiner* first transforms time interval-based data into endpoint-based data. Then, the problem of discovering temporal relationships is transformed into finding frequent sequences from endpoint-based data. Appliances' locations information is used to prune undesirable temporal relationships. The temporal distance among appliances is not taken into consideration, which can lead to many non-promising temporal relationships. For instance, the coffee maker is frequently used in the morning while the TV is frequently used in the evening. According to the study in [28], the two frequent appliances may be considered as temporally related. Actually, the two appliances are less likely to be temporally related even they occur frequently together. This is because their temporal distance is relatively large. In [30], the algorithm *CorFinder* is developed to find event correlations from sensor event data. The key idea of *CorFinder* is to find event correlations through transforming the problem of event correlation discovery into the known problem of frequent sequence discovery. A new graph-based method is presented to model the relations of the things based on a variety of information [29]. This information are extracted from things' usage time, location, and users' social network. After building the things graph, a Random Walk with Restart method is devised to find things relations. In our previous work [36], we propose an efficient algorithm *PMiner* to find spatio-temporal relationships among IoT services. A proximity metric is proposed to prune non-promising relationships among IoT services.

|  | taking medication | preparing breakfast | preparing lunch | preparing dinner | watching TV | preparing snacks | listening to music | toileting | washing dishes |
|---|---|---|---|---|---|---|---|---|---|
| taking medication | 67% | 11% | 0 | 21% | 0 | 11% | 0 | 0 | 0 |
| preparing breakfast | 0 | 90% | 10% | 0 | 0 | 0 | 0 | 0 | 0 |
| preparing lunch | 0 | 9% | 81% | 10% | 0 | 0 | 0 | 0 | 0 |
| preparing dinner | 0 | 0 | 0 | 92% | 0 | 8% | 0 | 0 | 0 |
| watching TV | 0 | 36% | 0 | 0 | 64% | 0 | 0 | 0 | 0 |

| preparing snacks | 7% | 20% | 9% | 21% | 0 | 43% | 0 | 0 | 0 |
|---|---|---|---|---|---|---|---|---|---|
| listening to music | 0 | 0 | 23% | 7% | 0 | 0 | 70% | 0 | 0 |
| toileting | 0 | 0 | 0 | 0 | 0 | 0 | 0 | 100% | 0 |
| washing dishes | 0 | 0 | 0 | 0 | 10% | 0 | 0 | 0 | 90% |

Fig. 11. Prediction accuracy

| | Day1 | Day2 | Day3 | Day4 | Day5 |
|---|---|---|---|---|---|
| Saved efforts | 8.4% | 10.3% | 14.7% | 6.4% | 6.5% |
| Saved time | 35min | 30min | 27min | 24min | 29min |

Fig. 12. Convenience

## 7.2   Temporal pattern mining

A temporal pattern is a set of time interval-based events that frequently occur together in particular temporal relations. Various temporal mining algorithms are designed based on the representation of temporal relationships. Allen's temporal model is widely employed to describe the temporal relationships among time interval events. In [31], an Apriori-based algorithm is proposed to discover temporal patterns. In [32], an arrangement numeration tree model is proposed. A hybrid DFS algorithm is designed based on the numeration tree data structure to search the temporal patterns. The hybrid DFS algorithm searches the temporal patterns by first transforming interval-based events into ID-lists and merging these Id-lists. In [33], a hierarchical representation augmented with addition information is introduced to model the temporal relationships. An efficient algorithm IEMiner is proposed which uses optimization strategies to reduce the candidate patterns generation. In [34], an efficient algorithm ARMADA is developed to search temporal patterns from large databases. This algorithm employs the pattern-growth principle from the frequent pattern mining area. Maximum time gap constraint is introduced to reduce non-promising candidate patterns. In [35], an end-point representation method is presented to represent the temporal relations among time interval events. The temporal pattern mining problem is transformed into sequential pattern mining. An efficient algorithm TPMiner is proposed to find temporal patterns. Pruning strategies are proposed to filter out undesirable temporal patterns.

## 7.3 Probabilistic sequential pattern

There are many research on the discovery of probabilistic sequential patterns considering uncertain and noisy features of sensor data. Various factors can contribute to data uncertainty such as data incompleteness, noise, and inaccurate measurements [37]. Introducing the probability concept is a promising solution to address the data uncertainty problem. In [37], a probabilistic frequency checking algorithm is developed to search the probabilistic frequent spatio-temporal sequential patterns considering time constraints. The patterns refer to a set of spatially close objects moving together for certain consecutive timestamps. In [38], an algorithm *Periodica* is designed to find the patterns of moving objects. The algorithm first finds the location of moving objects from trajectory data then finds the associated time intervals. An example of the patterns can be a person staying in the office from 9 am to 6pm. In [39], the concept of probabilistic sequential pattern is introduced in discovering surprising periodic patterns. An information gain measure is used to measure the surprise of the periodic patterns. In [40], an efficient algorithm is proposed to search long sequential patterns from noisy sensor data by introducing probability. In [41], the *UPref ixSpan* algorithm is developed to mine frequent sequential patterns from a large volume of noisy data by introducing probability concept to capture data uncertainty. The possible world semantics model is proposed to measure the pattern frequentness. Several pruning algorithms are designed to effectively solve the candidate generation explosion problem.

## 7.4 Activity recognition

A promising approach for recognizing human activity is to monitor and analyze human-object interactions. There are two main strands of approaches: knowledge-driven approach and data-driven approach [13]. The knowledge-driven approaches are based on the specification of activities by domain experts or users. The activities are generally modeled using involved objects and other characteristics of the activities such as users, location, and duration [9]. Ontology is one of the commonly used methods to represent activities [10]. Activities can be recognized by ontology reasoning based on available sensor data. Knowledge-driven approaches have two major advantages. The first is that these approaches do not require large datasets for training the activity model,
which is usually computing intensive. The second is that these approaches enable higher levels of automation via ontological reasoning [11]. However, the knowledge-driven approaches have limitations in handling noise and uncertainty in sensor data [12]. These approaches generally suppose that the residents' routine does not change thus they are limited in dealing with the variability of human activities in the long period of time.

Data-driven approaches are gaining more popularity for two reasons. On the one hand, sensor data regarding activity are generally noisy. On the other hand, occupants usually perform their daily activities in various manners [14]. There are different frameworks focus on facilitating sensor data to recognize human activities. One category of the data-driven approaches consider activity recognition as a supervised learning problem (i.e., classification problem). In the classification, the key idea is to extract features from sensor data, which are used to construct the classifier. Then the classifier is applied to predict the activity label for a sensor event sequence. In [15], a decision tree classifier is trained to recognition simple activities such as sitting and walking [14]. Naive Bayesian classifiers are used in [16] to recognize complex activities such as toileting and bathing from sensor data collected from daily objects. In [17], activities are represented as probabilistic event sequences and recognized from the interactions of daily objects. Other sophisticated supervised approaches include support vector machine [18], neural networks [19], and dynamic Bayesian network. When the temporal feature of activities is considered, the temporal classification approaches are developed. In temporal classification, probabilistic-based activity models are employed to recognize the hidden states (i.e., activity labels) from sensor data. The hidden Markov model (HMM) and the conditional random field (CRF) are among the most popular modeling approaches [20].The supervised approach is limited in its scalability because sensor data need to be annotated manually. An emerging pattern mining approach is proposed in [12] to recognize activities. Activities are modeled by emerging patterns which is a feature vector and can describe striking changes among different classes.

## 8 CONCLUSION

We proposed a novel Cognitive Amplifier framework to provide *quantifiable* convenience in terms of saving efforts and time in the smart home context. This framework consists of knowledge discovery and prediction components. The knowledge discovery component focuses on mining periodic probabilistic composition patterns and temporal relationships among composition patterns. The two types of knowledge serve as a base for building the prediction component. We conduct extensive experiments to demonstrate the *scalability* and *effectiveness* of the proposed algorithm. We also validate the applicability of the proposed approach. In our future work, we will extend the Cognitive Amplifier to include other sources of information such as the environment (e.g., temperature, sunlight, humidity) and semantic information (e.g., seasons). We will also investigate techniques that focus on multiple residents in the smart home settings.